\documentclass[10pt,twocolumn,letterpaper]{article}

\usepackage[pagenumbers]{iccv} 
\usepackage[accsupp]{axessibility}  

\definecolor{iccvblue}{rgb}{0.21,0.49,0.74}
\usepackage[pagebackref,breaklinks,colorlinks,allcolors=iccvblue]{hyperref}

\def\paperID{3} 
\def\confName{ICCV Workshop HiGen}
\def\confYear{2025}

\title{Controllable Pedestrian Video Editing for Multi-View Driving Scenarios \\ via Motion Sequence}

\author{Danzhen Fu$^{\star}$, Jiagao Hu$^{\star}$, Daiguo Zhou$^{\star}$, Fei Wang, Zepeng Wang, Wenhua Liao\\
MiLM Plus, Xiaomi Inc.\\
{\tt\small \{fudanzhen, hujiagao, zhoudaiguo, wangfei11, wangzepeng5, liaowenhua1\}@xiaomi.com}
}

\begin{document}
\maketitle

{\let\thefootnote \relax \footnote{$^{\star}$ Equal contribution.}}

\begin{abstract}

Pedestrian detection models in autonomous driving systems often lack robustness due to insufficient representation of dangerous pedestrian scenarios in training datasets. To address this limitation, we present a novel framework for controllable pedestrian video editing in multi-view driving scenarios by integrating video inpainting and human motion control techniques. Our approach begins by identifying pedestrian regions of interest across multiple camera views, expanding detection bounding boxes with a fixed ratio, and resizing and stitching these regions into a unified canvas while preserving cross-view spatial relationships. A binary mask is then applied to designate the editable area, within which pedestrian editing is guided by pose sequence control conditions. This enables flexible editing functionalities, including pedestrian insertion, replacement, and removal. Extensive experiments demonstrate that our framework achieves high-quality pedestrian editing with strong visual realism, spatiotemporal coherence, and cross-view consistency. These results establish the proposed method as a robust and versatile solution for multi-view pedestrian video generation, with broad potential for applications in data augmentation and scenario simulation in autonomous driving.

\end{abstract}    
\section{Introduction}
\label{sec:intro}

The rapid advancement of autonomous driving technology has been driven by breakthroughs in artificial intelligence algorithms and the accumulation of large-scale driving datasets. Among the critical components of perception systems, pedestrian detection models require extensive training data to handle complex real-world scenarios. However, the long-tailed distribution characteristic of driving data presents significant challenges, particularly for rare but safety-critical situations such as jaywalking or close-proximity pedestrians. These edge cases, though constituting a minimal proportion in natural driving data, demand robust model performance due to their high-risk nature - creating a paradox where data scarcity hinders the very capabilities that safety requirements necessitate. This underscores the importance of synthetic pedestrian data generation for autonomous driving systems.

Recent developments in diffusion-based video generation have enabled significant progress in controllable human motion synthesis. General-domain human motion generation models~\cite{wang2024disco, chang2023magicpose, xu2024magicanimate, karras2023dreampose, hu2024animate, zhu2024champ, wang2024unianimate, zhang2024mimicmotion} demonstrate remarkable capabilities in entertainment applications, where reference images combined with motion sequences from source videos allow generating characters performing specific actions. However, these single-view approaches cannot meet the multi-view requirements of autonomous driving systems, which typically employ multi-camera setups to capture 360° environmental contexts. This domain-specific requirement necessitates specialized solutions distinct from generic video generation frameworks.

In the autonomous driving domain, various multi-view video generation methods have been proposed~\cite{gao2024magicdrive3d, wang2024drivedreamer, zhao2025drivedreamer, zhao2025drivedreamer4d, ni2025recondreamer, gao2024vista, wen2024panacea}. For example, MagicDrive3D~\cite{gao2024magicdrive3d} and DriveDreamer4D~\cite{zhao2025drivedreamer4d} utilize 3D Gaussian Splatting~\cite{kerbl20233d} for novel view synthesis, while Vista~\cite{gao2024vista} integrates multi-modal control signals for future frame prediction. Panacea~\cite{wen2024panacea} adopts a two-stage pipeline that first generates multi-view images and subsequently uses them as initial frames for video generation. Although these methods achieve multi-view consistency in general driving scenes, they lack the ability to perform local pedestrian editing. To address this gap, we propose a dedicated framework for controllable pedestrian editing in multi-view driving scenarios.

In summary, We present a diffusion-based video editing framework that synergizes video inpainting with human motion sequence to achieve pedestrian manipulation in multi-view driving scenarios. Our principal contributions are threefold:

\begin{itemize}
    \item [1)] 
    \textbf{Local Video Editing Architecture}: We introduce a local cropping-and-generation paradigm that enables spatially-aware video synthesis through a dedicated video inpainting scheme.
    \item [2)]
    \textbf{Dynamic Pedestrian Region Cropping}: To tackle the scale-variation challenges in autonomous driving videos, we propose a two-stage adaptive processing pipeline: First, dynamically crop pedestrians according to their detection bounding box sizes, then resize the cropped regions to a uniform scale, enabling subsequent video generation to be performed consistently at this standardized scale.
    \item [3)]
    \textbf{Multi-View Consistency Framework}: We further propose a cross-view temporal-consistent video generation approach that ensures geometric alignment and semantic consistency across multi-camera perspectives.
\end{itemize}

Experimental results validate our framework's superiority in generating photorealistic pedestrian sequences with precise motion control across camera perspectives. These results demonstrate our approach as a robust solution for safety-critical pedestrian scenario generation, with potential applications in perception model augmentation and autonomous vehicle simulation systems.

\section{Related Works}
\label{sec:related}

\begin{figure*}[ht]
    \centering
    \includegraphics[width=\textwidth]{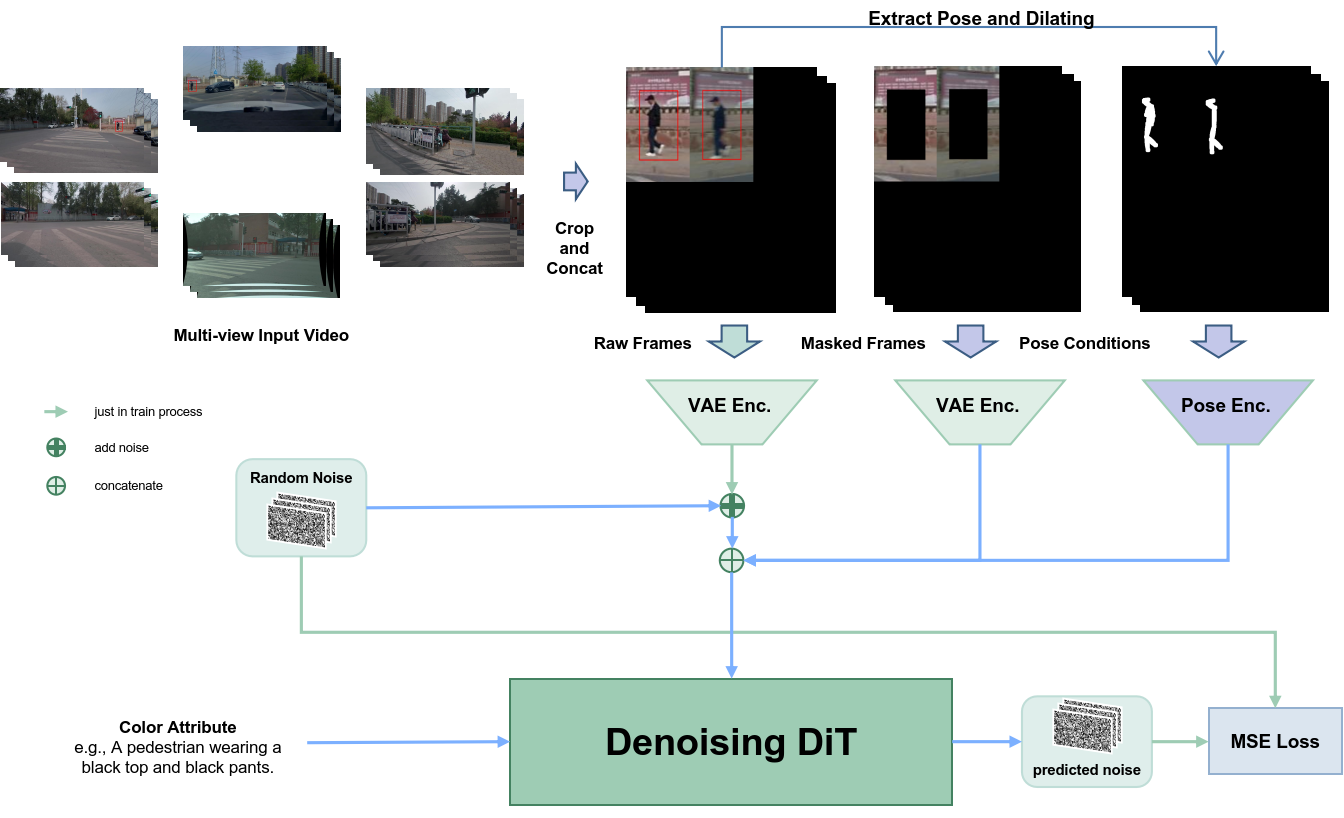}
    \caption{ Architecture of the proposed pedestrian editing pipeline with motion sequence control.}
    \label{fig:arc}
\end{figure*}

\subsection{Diffusion-Based Image Generation}

Diffusion models~\cite{nichol2021glide, rombach2022high, huang2023composer, saharia2022photorealistic} have emerged as the dominant paradigm in image generation, achieving state-of-the-art performance across various benchmarks. To address computational efficiency, Latent Diffusion Model~\cite{rombach2022high} introduces a two-stage framework that first encodes images into a compressed latent space, performs denoising in this lower-dimensional domain, and subsequently decodes to the pixel space - a strategy that balances generation quality and computational feasibility.

Particularly relevant to our work are conditional diffusion approaches that enhance controllability: ControlNet~\cite{zhang2023adding} and T2I-Adapter~\cite{mou2024t2i} introduce additional encoding pathways to preserve spatial constraints from edge maps, segmentation masks, and pose skeletons, enabling precise generation control through structural guidance. For identity-preserving generation, IP-Adapter~\cite{ye2023ip} employs cross-attention mechanisms to inject image prompts as reference conditions, while ObjectStitch~\cite{song2023objectstitch} utilizes CLIP~\cite{radford2021learning} embeddings to maintain semantic consistency during image composition. These works demonstrate the critical role of condition injection strategies in achieving both visual fidelity and controllability in diffusion-based frameworks.

\subsection{Diffusion-Based Video Generation}

The diffusion model's remarkable success in image generation has driven its extension to video synthesis. Many approaches~\cite{esser2023structure, khachatryan2023text2video, qi2023fatezero, singer2022make, wu2023tune} focus on enhancing temporal coherence while maintaining spatial quality through two primary strategies: motion-aware modules integration and spatiotemporal architecture redesign. Pioneering works like AnimateDiff~\cite{guo2023animatediff} demonstrate effective video generation by inserting motion modules into pre-trained text-to-image pipelines, leveraging inter-frame attention mechanisms to preserve temporal consistency without modifying base architectures. Meanwhile, recent advancements~\cite{zheng2024open, yang2024cogvideox, kong2024hunyuanvideo, wan2025wan} in native video diffusion models have generalized 2D denoising networks and variational autoencoders (VAEs) to 3D spatiotemporal domains. These 3D approaches achieve improved frame consistency but often require substantial computational resources.

\subsection{Human Motion Video Generation}
Recent advancements in pedestrian video generation have witnessed significant progress in controllable human motion synthesis. Early approaches primarily focused on text-driven generation, such as Text2Performer~\cite{jiang2023text2performer} which decoupled appearance descriptions from motion descriptions through dedicated language encoding. However, these purely language-based strategies often struggled to achieve precise control over both human appearance and motion dynamics. The subsequent Follow-Your-Pose~\cite{ma2024follow} framework introduced a two-stage paradigm, first generating base character images via text-to-image synthesis, then guiding motion generation using pose sequences.

Current research trends have shifted towards more sophisticated control mechanisms through pose sequence conditioning and reference image guidance~\cite{wang2024disco, chang2023magicpose, xu2024magicanimate, karras2023dreampose, hu2024animate, zhu2024champ, wang2024unianimate, zhang2024mimicmotion}. These methods leverage motion sequences for precise posture control while employing human reference images (refs) to encode appearance features, enabling person-specific animation generation. Notable approaches include Disco~\cite{wang2024disco}, MagicPose~\cite{chang2023magicpose}, MagicAnimate~\cite{xu2024magicanimate}, DreamPose~\cite{karras2023dreampose}, Animate Anyone~\cite{hu2024animate}, and Champ~\cite{zhu2024champ}, which integrate action sequence encoding for motion control with CLIP-based reference image encoding for appearance conditioning. Among these, Disco~\cite{wang2024disco} uniquely incorporates background encoding to facilitate scene replacement in generated videos. The Champ~\cite{zhu2024champ} framework further advances controllability by utilizing multi-modal motion sequences, including SMPL~\cite{loper2015smpl} (Skinned Multi-Person Linear) model representations that enhance 3D motion perception and produce more vivid, anatomically accurate animations.

The quality of motion guidance has also benefited from improved pose estimation techniques. Methods like Animate Anyone and Champ employ DWPose~\cite{yang2023effective} - an enhanced version of OpenPose~\cite{cao2017realtime} - which demonstrates superior accuracy in skeletal keypoint detection. This technical advancement contributes significantly to the generation of high-fidelity human motion videos with natural articulation and spatial coherence. The work presented in DensePose~\cite{guler2018densepose} aims to establish dense correspondences between an RGB image and a surface-based representation.

\section{Method}

Our proposed framework for controllable pedestrian generation in multi-view driving scenarios is illustrated in Figure~\ref{fig:arc}. The approach adopts a diffusion-based denoising paradigm, leveraging a ControlNet-like architecture to construct a video inpainting pipeline. Specifically, we first crop the target regions from the multi-view footage where pedestrians are to be inserted or replaced. The model then synthesizes pedestrian video sequences conditioned on predefined motion templates. Finally, the generated pedestrian content is seamlessly integrated into the original scenes across all views, enabling temporally coherent and spatially consistent pedestrian insertion and replacement.

\subsection{Dynamic Pedestrian Region Cropping}

\begin{figure}
    \centering
    \includegraphics[width=\linewidth]{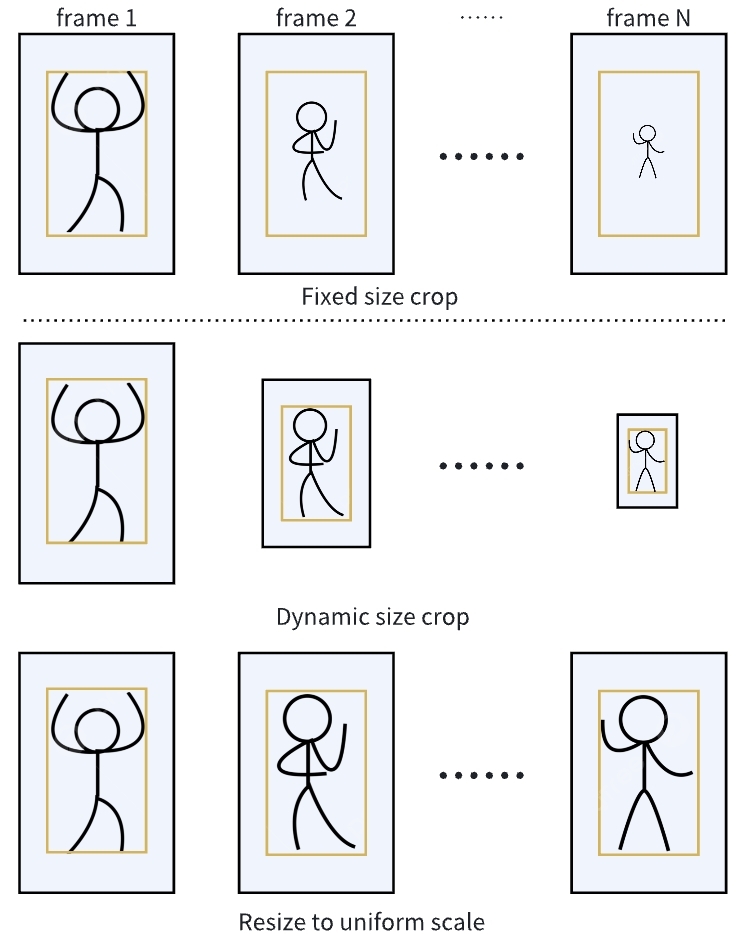}
    \caption{Dynamic-scale cropping with subsequent resizing to uniform scale}
    \label{fig:dynamic_crop}
\end{figure}

When fixed-size cropping was applied without dynamic adaptation, two critical visual artifacts emerged in multi-view driving scenarios. For close-range pedestrians, the rigid crop dimensions often failed to capture the entire body structure. This resulted in truncated limbs and incomplete poses, leading to visually implausible generations with distorted anatomical proportions or fragmented motion continuity. For distant pedestrians, the human body occupies an excessively small portion of the cropped frame, which significantly increases the difficulty of generation and hampers the recovery of fine-grained details. To overcome this limitation, we propose a dynamic pedestrian cropping strategy that standardizes pedestrian scales across views to simplify generation while enabling seamless multi-view integration. This dynamic crop-and-resize strategy is as illustrated in Figure~\ref{fig:dynamic_crop}. 

\subsection{Multi-View Consistency Framework}

We utilize multi-view image sequences annotated with LiDAR-based object detection. Given precise camera intrinsic and extrinsic calibration parameters, we first project 3D pedestrian annotations onto all camera views to obtain view-specific bounding boxes (bbox). These bbox coordinates are then expanded by a fixed margin ratio to capture contextual information and cropped pedestrian regions are resized to a uniform resolution (e.g., 480×240 pixels). For occluded or out-of-view pedestrians in specific camera angles, we employ zero-filled placeholder images to maintain view consistency. Finally, the processed multi-view pedestrian sequences are spatially organized into a composite image (e.g., arranged in a 2-row by 3-column layout) for unified generation.

\subsection{Video Inpainting}

We implement a standard video inpainting framework through mask-based conditioning, where the target pedestrians to be replaced or generated are masked out during training. The unmasked regions remain fully visible to the model and are encoded via VAE to produce control latents that preserve background context. These latents serve as critical constraints for background reconstruction while allowing pedestrian content manipulation.

Specifically, our masking protocol employs pedestrian bounding boxes as base masks, followed by spatial expansion and cropping operations. Each bounding box undergoes independent isotropic expansion along width and height dimensions, followed by region cropping that results in 1.6× the original bounding box dimensions. The expanded buffer zone facilitates seamless blending between generated pedestrians and background by maintaining visible transition areas.

\subsection{Pedestrian Motion Sequence Control}

Recent advancements in controllable video generation have demonstrated the effectiveness of using human pose sequences as conditional inputs for synthesizing realistic human motions. Notable frameworks such as Disco~\cite{wang2024disco}, AnimateAnyone~\cite{hu2024animate}, and Champ~\cite{zhu2024champ} have explored different strategies in this domain. AnimateAnyone~\cite{hu2024animate} employ 2D skeletal keypoint sequences as their primary control signals, while Champ innovatively combines four distinct pose representations - including depth images, normal maps, semantic maps obtained from SMPL sequences, and skeleton-based motion guidance -   through collaborative control to achieve superior detail preservation in generated videos.

In our approach, we adopt the minimalist skeletal keypoint representation for pedestrian motion control, motivated by two critical observations: First, in autonomous driving scenarios, pedestrians typically occupy only a small portion of the field of view in individual camera feeds, making fine-grained pose details less perceptible. Second, maintaining cross-view consistency across multi-camera systems proves more crucial than local detail fidelity when synthesizing traffic scenarios. To implement this, we utilize the open-source DWPose~\cite{yang2023effective} framework for keypoint extraction. For each camera view, we first perform cropping operations to isolate individual subjects before applying the pose estimation model to obtain temporally coherent keypoint sequences. These sequences are passed through a keypoint encoder to obtain motion latents, which are concatenated with inpainting latents from Section 3.2 to control pedestrian generation while maintaining background consistency.

This method follows standard denoising approaches: during training, VAE encodes videos into latents with added random Gaussian noise. The model predicts the noise through MSE loss minimization.
\section{Experimental Results}

\begin{figure*}
  \centering
  \includegraphics[width=\linewidth]{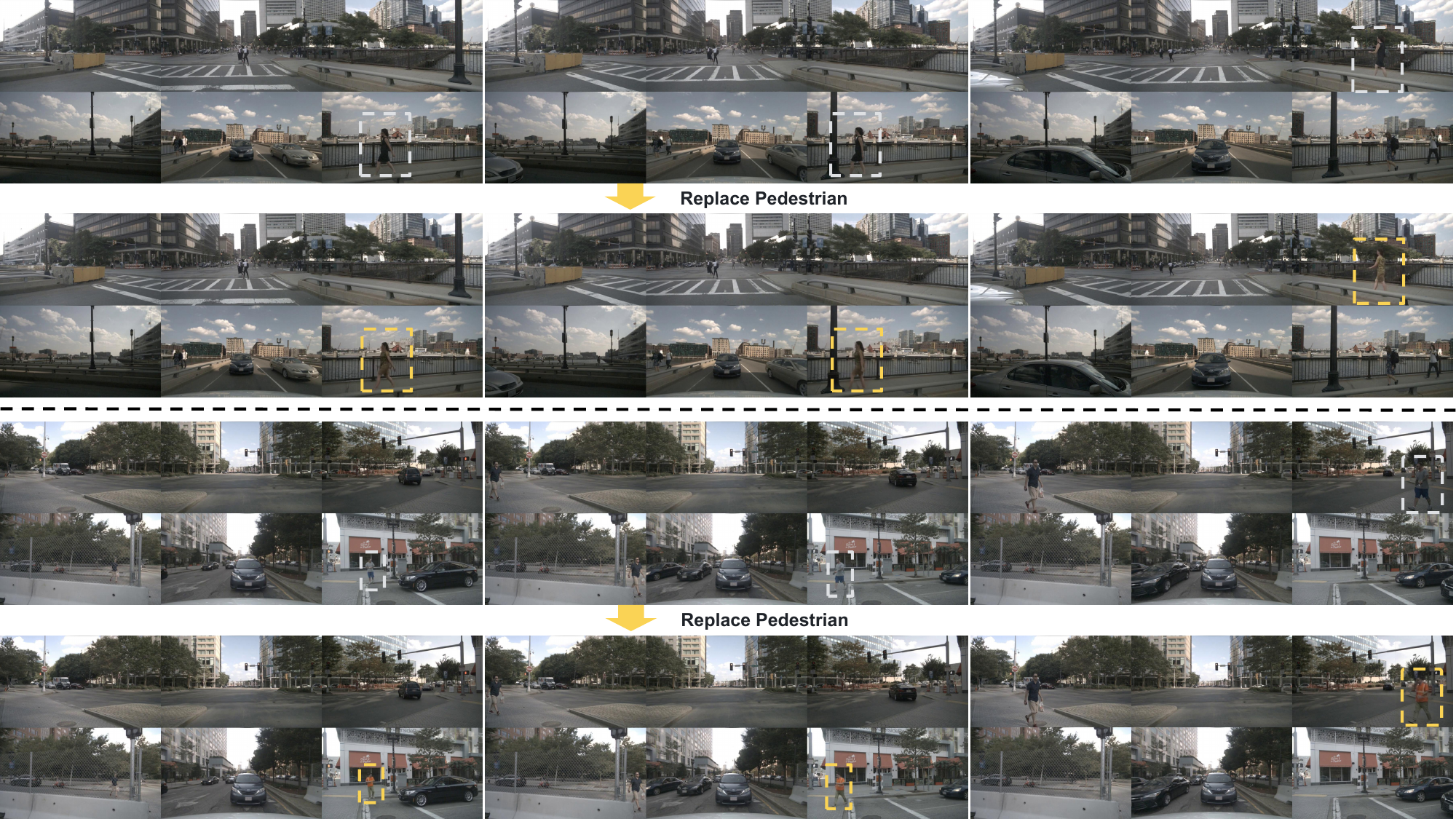}
  \caption{Qualitative results of pedestrian replacement in a multi-view scenario. Gray dashed boxes indicate the original pedestrian, while yellow dashed boxes highlight the generated pedestrian.
  }
  \label{fig:replace_pedestrian}
\end{figure*}

\subsection{Training Phase}

We trained the pedestrian video generation model on the nuScenes~\cite{caesar2020nuscenes} dataset using CogVideoX v1.5~\cite{yang2024cogvideox} as the base architecture. For single-view processing, pedestrian regions were cropped and resized to 480×240 resolution, while six-view inputs were tiled in a 2-row×3-column layout to form a composite resolution of (480×2)×(240×3). The model was trained on 85-frame sequences extracted from continuous video clips.

Motion cues were obtained through the DWpose estimator, followed by morphological dilation operations to enhance pose conditioning robustness. For view selection, we annotated the dominant viewpoint (most frequent occurrence) for each candidate pedestrian instance and localized their 2D bounding boxes. A pretrained video understanding model was then employed to extract clothing attributes, generating descriptions like ``a pedestrian wearing a white top and black pants" where only color terms (e.g., white/black) are replaceable within a fixed template structure.

During training, original videos were encoded into latent representations through the VAE module. The control conditions incorporated two components, Masked latent codes from the original video (with pedestrian regions masked out) and Encoded pose sequences. These were concatenated with noisy latent vectors following standard diffusion training protocols. The subsequent training procedures (noise prediction, optimization objectives) strictly followed the implementation of CogVideoX v1.5~\cite{yang2024cogvideox}.

\subsection{Inference Phase}

During inference, our framework first specifies a 3D motion sequence which is projected to multi-view perspectives using calibrated camera parameters to generate pose guidance maps. We then apply temporal-spatial masks to define editing regions and construct inpainting conditions. The pose maps and inpainting conditions are separately encoded via pose encoder and VAE encoder, then concatenated with noise latents in the DiT network architecture following training configurations. Text prompts (e.g., ``A pedestrian wearing a white top and black pants.") are converted into cross-attention embeddings through CLIP-based text encoder and integrated into DiT's cross-attention layers.

\subsection{Quantitative Assessment}

\begin{table}
    \begin{tabular}{c|cc}
    \toprule
    BEVFormer & w/o Synthetic Data  & w/ Synthetic Data \\ \midrule
    $AP_{dist\_0.5}$ & 0.1012 & 0.1063 \\
    $AP_{dist\_1}$ & 0.3552 & 0.3683 \\
    $AP_{dist\_2}$ & 0.5971 & 0.6136 \\
    $AP_{dist\_4}$ & 0.7173 & 0.7424 \\
    mAP & 0.4427 & \textbf{0.4577} \\
    \bottomrule
    \end{tabular}
    \caption{3D mAP metric for pedestrian detection.}
    \label{tab:compare}
\end{table}

\begin{figure*}
  \centering
  \includegraphics[width=\textwidth]{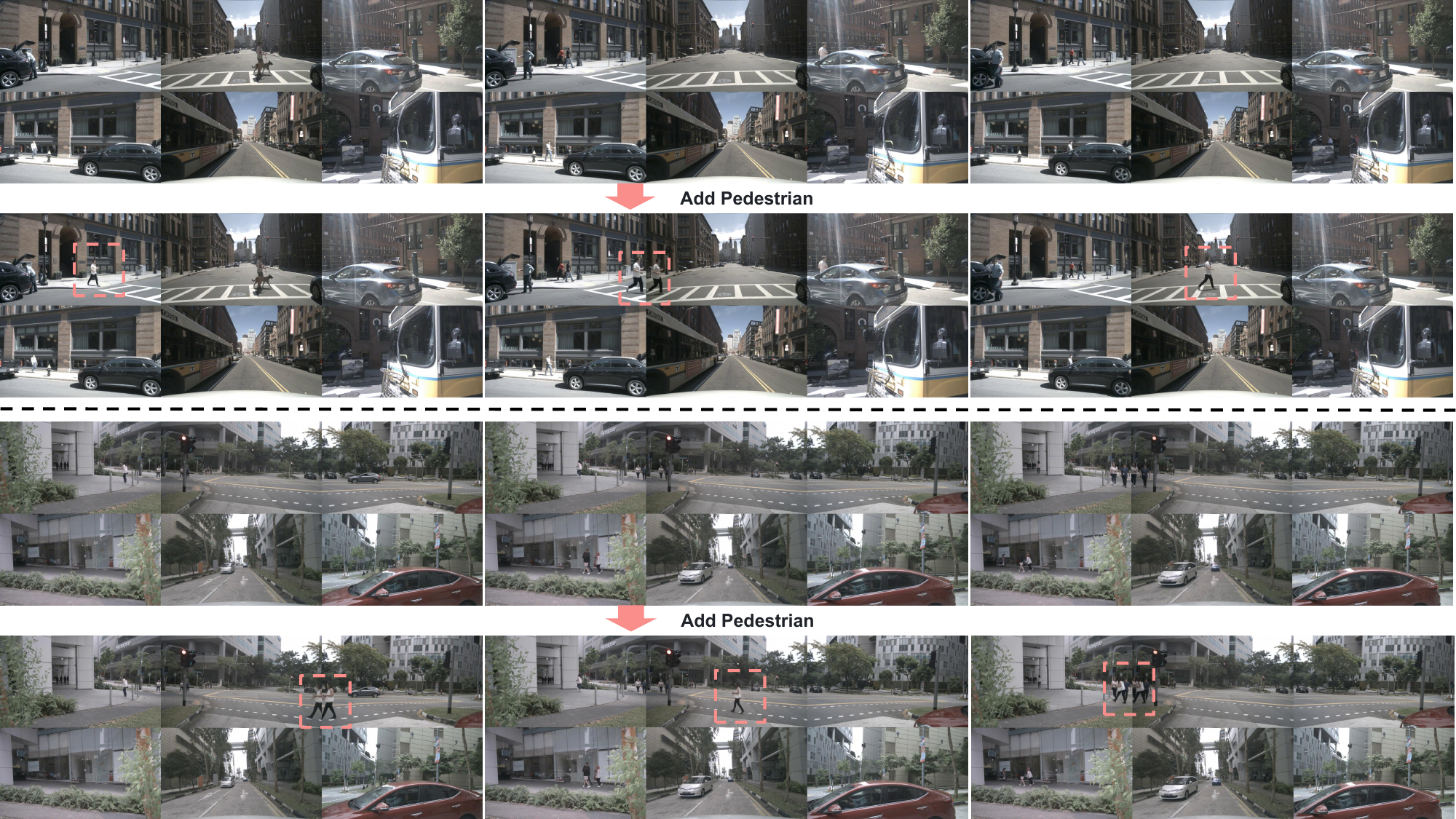}
  \caption{Qualitative results of pedestrian insertion in a multi-view scenario. Inserted pedestrians are highlighted by pink dashed boxes.}
  \label{fig:add_pedestrian}
\end{figure*}

We adopt the 3D mean Average Precision (mAP) as the primary evaluation metric, following the BEVFormer framework - a state-of-the-art multi-view perception model for autonomous driving. While mAP is a standard metric in 2D object detection tasks, its 3D counterpart employs distinct criteria for sample classification. Specifically, instead of using Intersection-over-Union (IoU) thresholds in image space, we utilize 2D center distance thresholds in Bird's Eye View (BEV) space, with standard thresholds set at {0.5m, 1m, 2m, 4m}. The final mAP score is obtained by averaging precisions across all distance thresholds.

To validate the effectiveness of our pedestrian generation approach, we conduct controlled experiments on the nuScenes dataset. The baseline model is trained on original data, while our proposed method replaces pedestrian instances in the training set with synthesized counterparts. Both models are evaluated on the nuScenes val set using mAP specifically for pedestrian class. Performance improvement in this metric would indicate that our motion-controllable pedestrian generation approach positively contributes to perception model training.

Table~\ref{tab:compare} presents the mAP evaluation results for the BEVFormerV2-R50-T1-Base model, a representative variant in the BEVFormer series. Compared to the baseline model trained on real-world data, our approach demonstrates significant performance gains in pedestrian detection when incorporating synthesized pedestrian samples for training. Specifically, the model trained with our generated data achieves notable improvements across all BEV distance thresholds. These results validate our method's capability in generating high-fidelity, multi-view pedestrian sequences that effectively enhance perception models' ability to detect pedestrians in complex driving scenarios.

The quantitative improvements in mAP scores provide empirical evidence that our motion-controllable pedestrian generation framework addresses critical challenges in multi-view detection tasks. By preserving spatial-temporal consistency while enabling motion sequence control, our synthesized data contributes to better generalization of the perception model.

\subsection{Qualitative Assessment}

\begin{figure*}
  \centering
  \includegraphics[width=\textwidth]{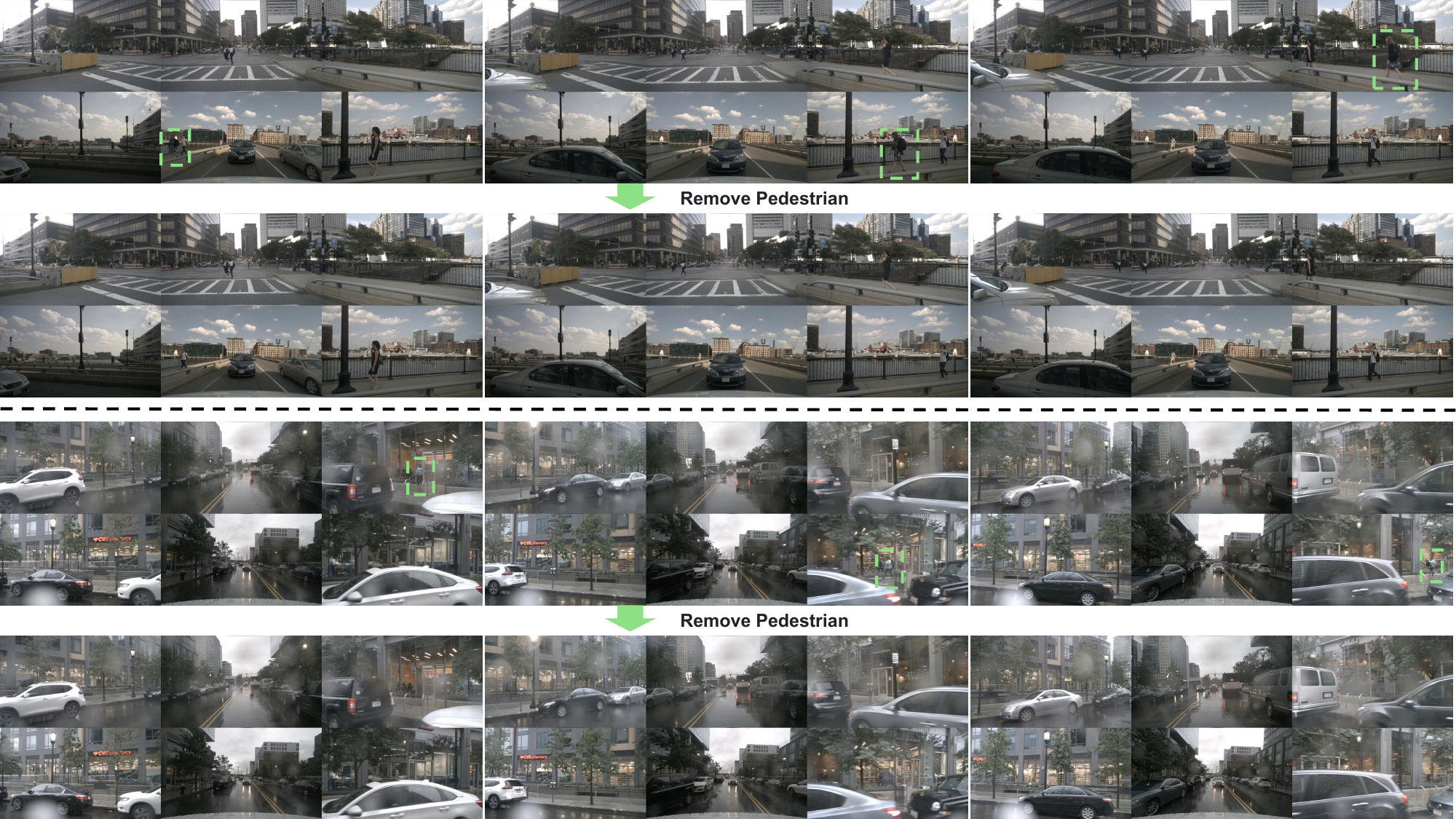}
  \caption{Qualitative results of pedestrian removal in a multi-view scenario. Pedestrians targeted for removal are indicated by green dashed boxes.}
  \label{fig:remove_pedestrian}
\end{figure*}

In this section, we qualitatively evaluate the capabilities of our method. Beyond modifying existing pedestrians, our framework also supports their insertion and removal, enabling a wide range of flexible editing operations within multi-view video sequences.

\paragraph{Pedestrian Replacement}


Figure~\ref{fig:replace_pedestrian} showcases our method's effectiveness in replacing pedestrians across synchronized multi-camera video streams. The generated pedestrians maintain consistent appearance across views and frames, and exhibit natural, temporally coherent motion. Furthermore, their behaviors are fully controllable via user-specified input conditions, enabling precise and customizable edits in complex, dynamic environments.

\paragraph{Pedestrian Insertion}


As illustrated in Figure~\ref{fig:add_pedestrian}, our approach enables the realistic insertion of synthetic pedestrians into vehicle-mounted multi-view footage. The inserted figures are visually indistinguishable from real pedestrians, and the synthesized content exhibits no observable artifacts or inconsistencies in either spatial or temporal dimensions. These results demonstrate strong generalization and highlight the model's ability to maintain coherence across multiple viewpoints.

\paragraph{Pedestrian Removal}


Pedestrian removal is achieved by setting the corresponding pose masks to zero, thereby suppressing the pedestrian signal while preserving the surrounding scene context. As shown in Figure~\ref{fig:remove_pedestrian}, the model successfully eliminates the specified pedestrians and reconstructs the occluded regions with high visual fidelity. The inpainted backgrounds blend seamlessly into the original scene, exhibiting no discernible discontinuities or visual artifacts.

\paragraph{Clothing Color Control}

\begin{figure*}
  \centering
  \includegraphics[width=\textwidth]{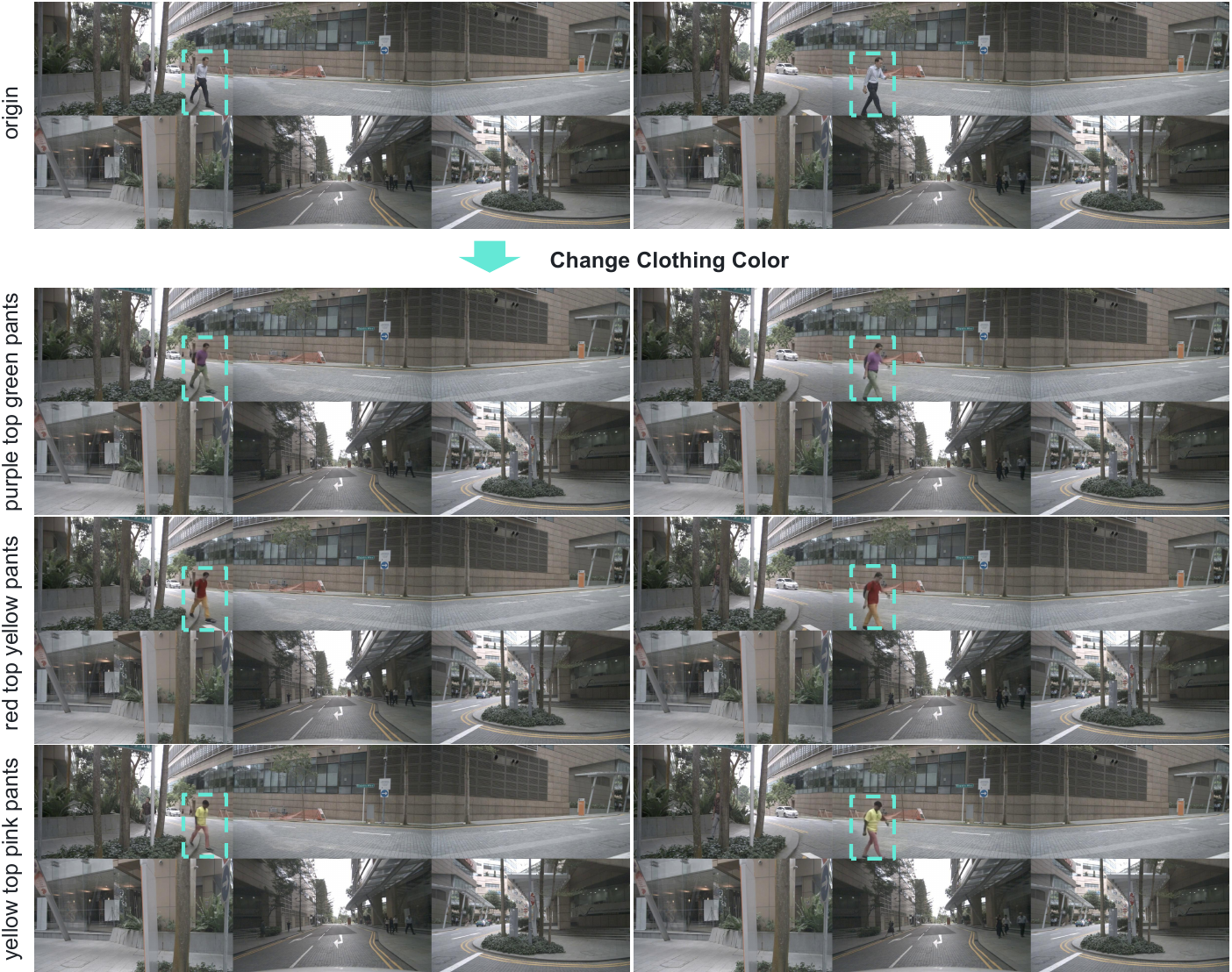}
  \caption{Qualitative results of clothing color manipulation in a multi-view scenario. Edited pedestrians are highlighted with cyan dashed boxes.}
  \label{fig:change_pedestrian}
\end{figure*}


Figure~\ref{fig:change_pedestrian} demonstrates the ability of our model to precisely control pedestrian clothing color through simple modifications in textual prompts. The generated results exhibit photo-realistic appearance and maintain contextual consistency across frames and views. This attribute-level control showcases the model's versatility in enabling fine-grained, text-driven visual edits while preserving both spatial alignment and temporal coherence.


\bigskip

In summary, these qualitative results highlight the robustness and flexibility of our method in executing comprehensive pedestrian editing tasks—ranging from replacement and insertion to removal and appearance customization. The edited sequences exhibit high perceptual quality, strong spatio-temporal consistency, and seamless integration within complex, multi-view scenes. These capabilities position our approach as a powerful and generalizable tool for controllable video editing in real-world applications.

\section{Conclusion}

This paper presents a novel multi-view pedestrian editing framework tailored for autonomous driving scenarios. Our approach synergistically integrates video inpainting and ControlNet-like architectural components to enable dynamic pedestrian synthesis through pose-guided cropping and stitching mechanisms. By leveraging clothing color prompts and pose sequence controls, the framework achieves comprehensive editability over pedestrian attire color and motion patterns in multi-view video generation. Notably, the proposed method ensures spatio-temporal consistency across all views while maintaining strict geometric alignment with the driving context.

Through empirical evaluation, we demonstrate the framework's effectiveness in generating photorealistic, multi-view consistent pedestrian sequences that preserve both appearance attributes and motion dynamics. The successful application of our synthesized data in BEVFormer model training highlights its practical value in expanding limited real-world pedestrian samples for autonomous driving perception tasks. Experimental results show that the generated data improves detection performance. This validates our method's capability to produce high-quality, spatial-temporal consistency pedestrian samples that faithfully reflect real-world driving contexts.

{
    \small
    \bibliographystyle{ieeenat_fullname}
    \bibliography{main}
}

\end{document}